\documentclass[letterpaper, 10pt, conference]{ieeeconf}
\pdfoutput=1
\IEEEoverridecommandlockouts 
\usepackage{import}
\usepackage{style}

\makeatletter
\let\NAT@parse\undefined
\makeatother
\usepackage[pdfborder={0 0 1}]{hyperref}  

\title{\LARGE \bf
Online Trajectory Planning Through Combined Trajectory Optimization and Function Approximation: Application to the Exoskeleton Atalante
}

\author{Alexis Duburcq$^{1,2,3}$, Yann Chevaleyre$^{2}$, Nicolas Bredeche$^{3}$ and Guilhem Bo\'{e}ris$^{1}$%
\thanks{$^{1}$Wandercraft, Paris, France. \href{mailto:alexis.duburcq@dauphine.eu}{\nolinkurl{<alexis.duburcq@dauphine.eu>}}.
}%
\thanks{\hbadness=2100 $^{2}$Universit\'{e} Paris-Dauphine, PSL, CNRS, Laboratoire d'analyse et modélisation de systèmes pour l'aide à la décision, Paris, France.
}%
\thanks{$^{3}$Sorbonne Universit\'{e}, CNRS, Institut des Syst\`{e}mes Intelligents et de Robotique, ISIR, F-75005 Paris, France.}%
}

\begin{document}

\clubpenalty=9996
\widowpenalty=9999 
\brokenpenalty=4991
\predisplaypenalty=10000
\postdisplaypenalty=1549
\displaywidowpenalty=1602

\maketitle


\begin{abstract}

Autonomous robots require online trajectory planning capability to operate in the real world. Efficient offline trajectory planning methods already exist, but are computationally demanding, preventing their use online. In this paper, we present a novel algorithm called Guided Trajectory Learning that learns a function approximation of solutions computed through trajectory optimization while ensuring accurate and reliable predictions. This function approximation is then used online to generate trajectories. This algorithm is designed to be easy to implement, and practical since it does not require massive computing power. It is readily applicable to any robotics systems and effortless to set up on real hardware since robust control strategies are usually already available. We demonstrate the computational performance of our algorithm on flat-foot walking with the self-balanced exoskeleton Atalante.

\end{abstract}


\section{INTRODUCTION}

Online trajectory planning enables robots to deal with a real-world environment that may change suddenly, and to carry out sequences of tasks in unknown orders and contexts. For instance, walking robots must be able to change direction, adapts their speed, consider stairs of different heights, or the position and size of obstacles.  Yet offline trajectory planning remains very challenging for complex systems that may involve hybrid dynamics, underactuation, redundancies, balancing issues, or a need for high accuracy. Although methods exist to solve most trajectory optimization problems, there is no guarantee of convergence and finding solutions is computationally demanding, preventing their uses online. \par
One way to get around these issues is running the optimization in background and updating the trajectory periodically, e.g. between each step for biped robots \cite{Hereid2016b}. However, it remains hard to meet such computational performance, and this still provides a poor reaction time. Another approach is to simplify the model to speed up the calculation and ensure convergence, for example by linearizing it. Nevertheless, it does not have any guarantee to be feasible in practice since it does not take into account the actual dynamics of the system, and the overall motion is less sophisticated \cite{De2012, Kajita2003}. \par \medskip
A workaround to avoid online trajectory optimization consists in using a function approximation, ie. to perform trajectory learning over a set of trajectories optimized beforehand. This requires no simplification of the model since the optimizations are carried out offline. Moreover, once training has been done, it operates at a fraction of the cost of the previous methods. Two distinct approaches can be considered: policy learning, i.e training a controller, and trajectory learning, i.e. predicting nominal state sequences. This paper focuses on trajectory learning. Indeed, it has the advantage of being effortless to implement on robotic systems for which there already exists control strategies that ensures robust tracking of trajectories generated through optimization: it comes down to replacing a finite set of trajectories by the function approximation. \par

\begin{figure}[t] \centering 
\subfloat{
    \includegraphics[width=.34\linewidth, valign=c]{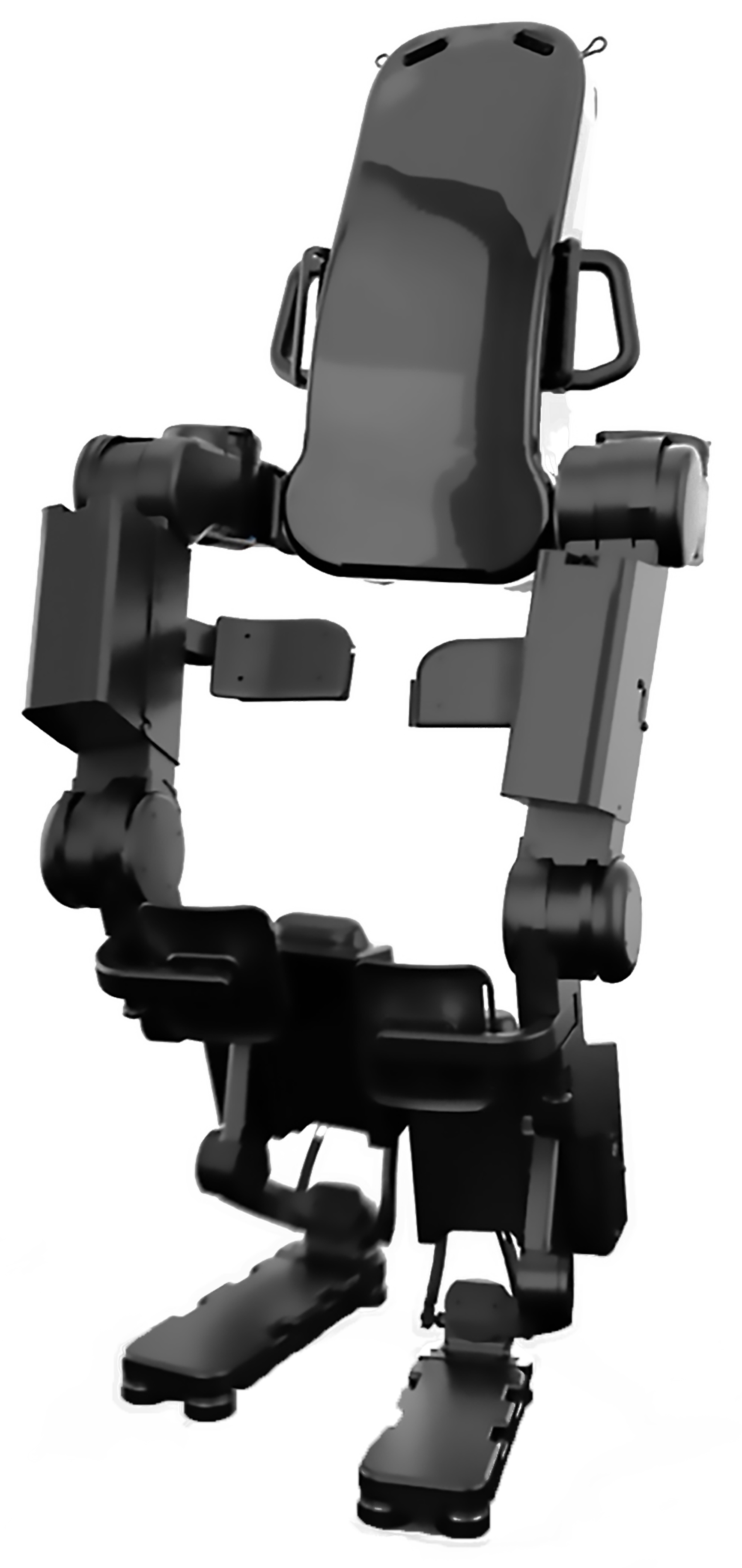}
} 
\hspace{2.5em}
\subfloat{
    \includegraphics[width=.296\linewidth, valign=c]{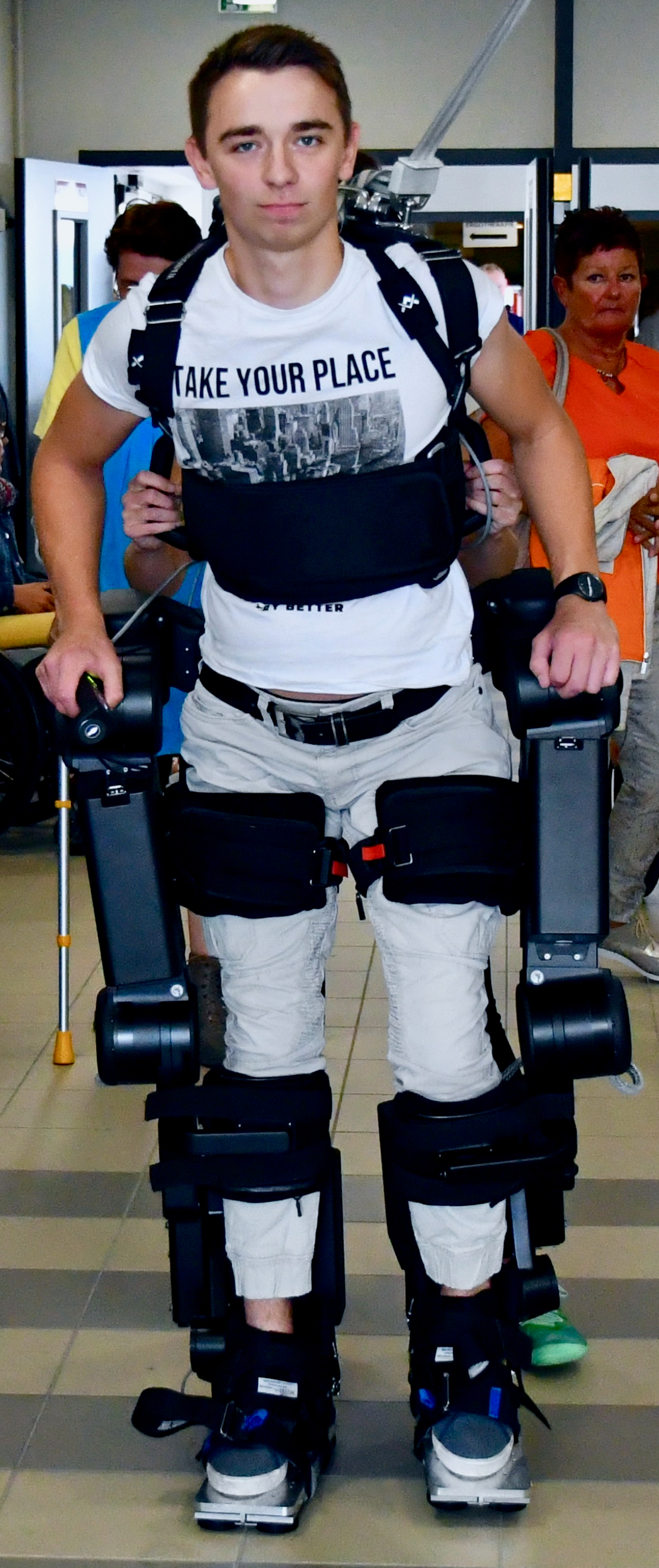}
}
    \vspace{0.5em}
    \caption{Rendering of the exoskeleton Atalante on the left. Clinical trials assisted by a physiotherapist for safety on the right.}
    \label{fig:exoskeleton_mdl} \vspace{-1.4em}
\end{figure}
A naive approach would be to train a function approximation on a database of solutions of the optimization problem. Although it may work in practice, this does not offer any guarantee to really perform the desired task nor to be feasible, and is sensitive to overfitting. This is actually the state-of-the-art in trajectory learning, as this field is still largely unexplored. On the contrary, there has been major advances in policy learning. It is now possible to ensure accurate and reliable learning, which is exactly what trajectory learning is not able to do yet. Still, it can hardly be applied on robot hardware because the reality gap is not properly handled. \par \medskip
Our contribution is the Guided Trajectory Learning algorithm (GTL), which makes trajectory optimization adapt itself, so that it only outputs solutions that can be perfectly represented
by a given function approximation. It is inspired by the Guided Policy Search (GPS) \cite{Levine2013} and the work of Mordatch \cite{Mordatch2014} in the context of policy learning. The idea is to make the trajectory optimization problem adapt itself wherever the function approximation fails to fit. Adaptation is achieved by jointly solving trajectory optimization and trajectory learning for a collection of randomly sample tasks. This is a consensus optimization problem over a database of trajectories, which is intractable directly. We overcome this limitation by solving it iteratively via the Alternating Direction Method of Multipliers (ADMM) \cite{Boyd2011}. Furthermore, as function approximation, we propose a special kind of deconvolution neural network well-suited for fitting trajectories. \par \medskip
Our method is readily applicable to any complex robotics systems with high-dimensional state for which offline trajectory optimization methods and robust control strategies are already available and efficient. It makes online trajectory planning based on function approximations more accurate and reliable by guaranteeing the feasibility of the predictions, and thereby it is practical for systems where failure is not an option. We demonstrate it on flat-foot walking with the medical exoskeleton Atalante designed by Wandercraft \cite{Wandercraft}.
\section{RELATED WORK}

Even though policy learning may exhibit richer and nicer behaviors than trajectory learning since it performs feedback control, it comes with additional challenges. Indeed, the performance of the few policy learning approaches assessed on real robots were unsatisfactory because of the reality gap \cite{Hays2012}. Although major advances were made recently \cite{Bousmalis2018}, it is not near to be readily available for systems already challenging to control using classic methods. Daumé III suggests to get around this problem by using policy learning to predict trajectories \cite{Daume2009}. This can be done by simply passing on earlier predictions as inputs for future predictions. Although trajectories computed in this way can achieve results competitive with trajectory learning \cite{Ross2010}, it makes policy learning an indirect approach to do trajectory learning, therefore losing its main advantages wrt. trajectory learning. \par \medskip
Both policy and trajectory learning are prone to the so-called distributional shift issue, namely the predictions themselves affect the future states during the execution on the system. Ignoring this leads to very poor performance in the case of policy learning \cite{Ross2010, Bagnell2010, Abbeel2016}. This is because as soon as the policy makes a mistake, it may encounter completely different observations than those under expert demonstrations, leading to the compounding of errors. The Guided Policy Search (GPS) \cite{Levine2013, Levine2013a} overcomes this limitation by adapting the states for which expert demonstrations are provided, but also the policy optimization problem itself. Formally, the optimization problem is modified to maximize the actual return of the learned policy, making it less vulnerable to suboptimal experts. The resulting policy is guaranteed to perform well under its induced distribution of states. \par
The effect of the distributional shift is less dramatic in trajectory learning, since the observations do not take part in the prediction process. Its effect follows directly from the reconstruction error of the function approximation, and thereby requires the latter to fit accurately. Even though trajectory learning using a standard regression may be appropriate in some cases, this does ensure accurate predictions. This approach has proven effective on the bipedal robot Cassie for controlling in real-time the velocity of the center of mass \cite{Da2017, Xie2018}, but was unsuccessful in simulation for more complex systems like a human-sized 3D humanoid \cite{Levine2013}. \par \medskip
Unlike GPS, the policy learning method proposed by Mordatch and Todorov \cite{Mordatch2014} is closely related to trajectory learning since it relies on trajectory optimization to generate expert demonstrations. They seek after a compromise between the optimality of the trajectories and the accuracy of the learned policy, and solves this problem via the Alternating Direction Method of Multipliers. Robustness of the feedback control loop is improved by training the policy to behave as a Linear Quadratic Regular \cite{Kwakernaak1972}. However, their formulation of the problem does not allow to cancel the reconstruction error out and therefore to get rid of the distributional shift completely. Although this approach is cost-efficient and leads to satisfactory results in simulation for simple systems, it is likely to diverge in practice. \par \medskip
Another aspect to consider is that trajectory optimization based on the direct collocation framework is numerically robust and scalable \cite{Kelly2017, Hargraves1987}. It has proven its ability to efficiently solve most trajectory optimization problems, even for complex robotic systems, such as humanoid robots walking trajectory generation \cite{Hereid2016, Hereid2017, Grizzle2018}. Therefore, adapting the policy learning approaches to trajectory learning appears natural to overcome their respective limitations. 
\section{PRELIMINARIES}

\subsection{Trajectory Optimization Problem}

Consider a time-invariant time-continuous dynamical system of the form $\dot{x}(t) = f(x(t),u(t))$, where $x(t) \in \mathbb{R}^{p},\, u(t) \in \mathbb{R}^q$ are the state and the controls of the system applied at time $t$, respectively. $f$ denotes the dynamics of the system. Optimization variables minimizing a cost function are highlighted with a superscript asterisk $\cdot^*$. \par \medskip
Given a task to perform $\tau \in \mathcal{D}_\tau$, where $\mathcal{D}_\tau$ is a compact set of $\mathbb{R}^m$ denoting the task space, a trajectory optimization problem for such a system can be formulated as 
\begin{equation} \label{eq:traj_opt}
    (x^*, u^*, t^*) = \underset{(x,u,T) \in \mathcal{C}_\tau}{\argmin}\mkern+05mu \int_0^T l(x(t),u(t),T) \, dt,
\end{equation}
where $x:t \mapsto x(t), u:t \mapsto u(t)$ are functions whose temporal dependence is implicit, $l$ is the running cost, and $T$ is the duration of the trajectory. $\mathcal{C}_\tau$ is the feasibility set
\begin{align*}
    \mathcal{C}_\tau = \{
    &(x,u,T) \in \mathcal{D}_x \times \mathcal{D}_u \times \mathcal{D}_T \mid \\ & c_{T}\left(\tau,x(0),u(0),x(T),u(T),T\right) = 0, \\
    & c_{in}(\tau,x(t),u(t),t) \leq 0, \\
    & c_{eq}(\tau,x(t),u(t),t) = 0\},
\end{align*} \\[-1.5em]
where $\mathcal{D}_x, \mathcal{D}_u, \mathcal{D}_T$ are compact sets embodying the physical limitations of the system. The terminal constraints $c_T$ and the inequality and equality constraints $c_{in}, c_{eq}$ are continuously differentiable functions that depends on the task $\tau$. \par \medskip
The periodicity and duration of the trajectory are examples of terminal constraints, while the dynamics equation and the admissibility conditions are part of the inequality and equality constraints. A task can be composed of any combination of high-level objectives: for example, for a walking robot, the desired step length and speed. \looseness=-1 \par \medskip
The state and control functions $x, u$ are further discretized in time sequences of fixed length $L_T$. In this context, the optimal trajectory for task $\tau$ is uniquely defined by its sequence of states and duration $(\{x^*_1, x^*_2, \dots, x^*_{L_T}\}, T^*)$.

\subsection{Trajectory Learning Problem}

The objective is to use the solutions generated through trajectory optimization to train a function approximation parametrized by $W \in \mathbb{R}^n$, such that, for any task $\tau$, it outputs a trajectory achieving the task. As the control strategies are not part of the learning process, this can be viewed as a standard regression over a database of $N$ optimal trajectories $\{\tau_i,(X^*_i=\{x^*_{i,1}, x^*_{i,2}, \dots, x^*_{i, L_T}\}, T^*_i)\}_{i = 1}^N$, where $N$ must be sufficiently large to span the whole task space. \par \medskip
Formally, training a function approximation consists in finding parameter $W$ giving the best performance in average, \vspace{-0.8em}
\begin{equation} \label{eq:reg_pb}
    W^* = \underset{W \in \mathbb{R}^n}{\argmin}\mkern+05mu R_\gamma(X^*,T^*,W), \\[-0.2em]
\end{equation} \\[-0.7em]
where $R_\gamma(X,T,W)$ is the total reconstruction error, such that \vspace{-0.7em}
\begin{equation} \label{eq:reg_err} \hspace{-0.55em}
    R_\gamma(X,T,W) \triangleq\! \sum_{i=1}^{N} \left\| X_i - \hat{X}(\tau_i,W)) \right\|^2 \!+ \gamma \left\| T_i - \hat{T}(\tau_i,W) \right\|^2 \!\!,
\end{equation} \\[-0.8em]
where $\gamma$ is a weighting factor that determines the trade-off between the state and duration fitting accuracy. The predicted duration and state sequence for a task $\tau$ are denoted $\hat{T}(\tau,W), \hat{X}(\tau,W)$. The subscript $\cdot_i$ specifies the task. \par \medskip
It is irrelevant to take into account the constraints of the trajectory optimization problem explicitly in the regression since they are satisfied at the limit when the reconstruction error vanishes. Reducing this error is usually done by increasing number of fitting parameters $n$ to match the regularity of the data. However, it does not bring any guarantee regarding the feasibility of the predictions apart from the training samples, since overfitting may occur. Several techniques exist to alleviate this issue without increasing the computational cost too much, e.g. early stopping and regularization \cite{Girosi1995, Srivastava2014, Zou2005}, but they do not allow adaptation of the training data. Regularisation of the training data through adaptation may be more computationally demanding, but it ensures reliable predictions by limiting the number of fitting parameters.
\section{GUIDED TRAJECTORY LEARNING}

We propose to achieve adaptation of the training data by solving simultaneously trajectory optimization and trajectory learning, such that the generated trajectories are perfectly fitted by the function approximation and satisfy the constraints of the original trajectory optimization problem concurrently.

\subsection{Consensus Optimization Problem}

This problem consists of jointly optimizing a collection of $N$ trajectories achieving a set of tasks $\{\tau_i\}_{i \leq N}$, and the parameters of the function approximation $W$ that is intended to fit them. This can be written as follows
\begin{equation} \label{eq:joint_pb} \hspace{-0.55em}
    (X^*, U^*, T^*, W^*) \, =\mkern-15mu \underset{\substack{
        (X_i,U_i,T_i)_{1 \leq i \leq N} \, \in \, \prod_{i=1}^N \mathcal{C}_{\tau_i} \\[0.1em]
        W \in \mathbb{R}^n}
    }{\argmin}\mkern+3mu  \frac{1}{N} \sum_{i=1}^N L(X_i,U_i,T_i)
\end{equation} \vspace{-0.7em}
\begin{equation*}
    \text{st. } X_i = \hat{X}(\tau_i,W), T_i = \hat{T}(\tau_i,W), \, \forall i \in \{0,1,\dots,N\}, \vspace{-0.3em}
\end{equation*}
where the tasks $\{\tau_i\}_{i \leq N}$ are uniformly sampled, $L(X_i,U_i,T_i)$ is the discretization of the total optimization cost \eqref{eq:traj_opt} \vspace{-0.3em}
\begin{equation*}
    L(X_i,U_i,T_i) \triangleq \frac{T_i}{L_T} \sum_{t = 1}^{L_T} l(x_{i,t},u_{i,t},T_i).
\end{equation*} \par \vspace{-0.3em}
The reconstruction constraint guarantees that the trajectories are perfectly fitted by the function approximation, despite its potentially limited expressive power, and have properties compliant with it. However, solving this problem directly is intractable since it requires to compute the gradient of every trajectory optimization sub-problems at every step of the solver. Indeed, the number of sub-problems $N$ usually ranges from thousands to hundreds of thousands.

\subsection{Alternating Direction Method of Multipliers}

ADMM \cite{Boyd2011} is a method to efficiently solve optimization problems composed of a collection of subproblems linked by a single linear equality constraint but otherwise independent, each of them having a readily available solving method. Let's consider the following separable nonconvex consensus problem \cite{Magnusson2016, Andreani2007} \vspace{-0.4em}
\begin{equation} \label{eq:joint_pb_gen}
    (Y^*,Z^*) = \mkern-15mu \underset{\substack{
        (y_i)_{1 \leq i \leq N} \, \in \, \prod_{i=1}^N \mathcal{Y}_i \\[0.2em]
        Z \in \mathcal{Z}}
    }{\argmin}\mkern+05mu \sum_{i=1}^N f_i(y_i) + g(Z)
\end{equation} \vspace{-1.5em}
\begingroup
\setlength{\belowdisplayskip}{0.4em}
\begin{align*}
    \text{st. } & \resizebox{0.95\hsize}{!}{$\mathcal{Y}_i = \left\{y_i \in \mathcal{D}_y \mid \psi_i(y_i) = 0, ~ \phi_i(y_i) \leq 0\right\}, \forall i \in \{0,1,\dots,N\}$} \\
    & \mathcal{Z} = \left\{Z \in \mathcal{D}_z \mid \theta(Z) = 0, ~ \sigma(Z) \leq 0\right\} \\
    & Y - Z = 0,
\end{align*}
\endgroup
where $\mathcal{D}_y \subset \mathbb{R}^p, \mathcal{D}_z \subset \mathbb{R}^{N p}$ are compact sets. The cost functions $f_i:\mathbb{R}^{p} \to \mathbb{R}, g:\mathbb{R}^{N p} \to \mathbb{R}$ and the equality and inequality constraints $\psi_i, \phi_i, \theta, \sigma$ are twice continuously differentiable. \par \medskip
Such nonconvex problems can be handled efficiently by the Augmented Lagrangian Method \cite{Bertsekas1976, Bertsekas1982}. In this case, the augmented Lagrangian in scaled form can be stated as \vspace{-0.2em}
\begingroup
\setlength{\belowdisplayskip}{0.5em}
\begin{equation*}
    L_\rho(Y,Z,\Lambda) \triangleq \sum_{i=1}^N f_i(y_i) + g(Z) + \frac{\rho}{2} \left\| Y - Z + \Lambda \right\|^2.
\end{equation*}
\endgroup
This is an exact penalty method \cite{Han1979, Di1989}. In this regard, the original problem can be solved by minimizing it. This can be done in an alternating Gauss-Seidel manner, optimizing each variable while holding the others fixed \cite{Boyd2011, Bezdek2003}. This yields Algorithm \ref{admm_algo} taken from \cite{Magnusson2016, Andreani2007}. $\alpha$ and $\rho^k$ are referred to as the dual step size and penalty factor, respectively.

\subsection{Guided Trajectory Learning}

Let us introduce an additional optimization variable $Z$ in Problem \eqref{eq:joint_pb}, such that $Z_i \triangleq (\hat{X}(\tau_i,W), \gamma \hat{T}(\tau_i,W))$, where $(\cdot,\cdot)$ stands for the vector concatenation operator. \par \vspace{1.0em}

\begingroup \vspace*{-0.8em}
\removelatexerror
\SetAlgoSkip{}
\LinesNumberedHidden
\SetAlCapHSkip{0.3em}
\setlength{\algowidth}{0.2\hsize}
\begin{algorithm}[H] \label{admm_algo}
    \Numberline initialization\;
    \While{stopping criterion not met}{
        \Indpp
        \For{$i\in\{1,...,N\}$}{ \vspace{-0.3em}
            \begin{flalign*}
                \Numberline y_i^{k+1} &= \underset{y_i \in \mathcal{Y}_i}{\argmin}\mkern+05mu f_i(y_i) + \frac{\rho^{k}}{2} \left\| y_i - z_i^{k} + \lambda^{k}_i \right\|^2 &&
            \end{flalign*}
        } \vspace{-1.5em}
        \begin{flalign*}
            \Numberline Z^{k+1} &= \underset{Z \in \mathcal{Z}}{\argmin}\mkern+05mu g(Z) + \frac{\rho^{k}}{2} \left\| Y^{k+1} - Z  + \Lambda^{k} \right\|^2  && \\[5pt]
            \Numberline \Lambda^{k+1} &= \Lambda^{k} + \alpha (Y^{k+1} - Z^{k+1}) \label{eq:admm_multi} &&
        \end{flalign*} \vspace{-1.35em}
    }
    \caption{ADMM for Nonconvex Consensus Problem}
\end{algorithm} \vspace{1.4em}
\endgroup

The reconstruction constraints becomes
\begin{gather*}
    (X_i,\gamma T_i) - Z_i = 0 \\[0.2em]
    \text{st. } Z_i \in \mathcal{Z}_\gamma = \{z \mid \underset{W}{\inf}\mkern+05mu \left\| z - (\hat{X}(\tau_i,W),\gamma \hat{T}(\tau_i,W)) \right\|^2 = 0 \}.
\end{gather*}

Then, ADMM can be applied to solve Problem \eqref{eq:joint_pb}. The \par
\noindent Augmented Lagrangian is given by
\begin{equation*}
    L_\rho(X,U,T,Z,\Lambda) = \frac{1}{N}\!\sum_{i=1}^N L(X_i,U_i,T_i) + \frac{\rho}{2} \left\| (X,\gamma T) - Z + \Lambda \right\|^2 \vspace{-0.2em}\!\!\!,
\end{equation*}
where $\Lambda=(\Lambda_X,\gamma \Lambda_T)$. \par \medskip

\noindent The update rule for step 2 of Algorithm \eqref{admm_algo} corresponds to
\begin{multline} \label{eq:admm_opt}
    (X_i^{k+1}, \, U_i^{k+1}, \, T_i^{k+1}) =  \\
    \underset{(X_i,U_i,T_i) \in  \mathcal{C}_{\tau_i}}{\argmin} L(X_i,U_i,T_i)
    + \frac{\rho^k}{2} \left\| (X_i,\gamma T_i) - Z_i^k + \lambda_i^k \right\|^2\!\!,
\end{multline}

\noindent while the update rule for step 3 is
\begin{equation} \label{eq:admm_reg}
\begin{gathered}
    Z_i^{k+1} = (\hat{X}(\tau_i,W^{k+1}), \gamma \hat{T}(\tau_i,W^{k+1})) \\[0.2em]
    \text{st. } W^{k+1} = \underset{W \in \mathbb{R}^n}{\argmin}\mkern+05mu R_\gamma(X^{k}+\Lambda_X,T^{k}+\Lambda_T,W).
\end{gathered}
\end{equation}

One can think of the multipliers $\Lambda$ being the cumulative residual prediction error for each task $\tau_i$. They reveal where the function approximation makes repeating prediction errors for each trajectory. They modify the regression and trajectory optimization objective functions to give more weight to regions where errors are consistently made. Over iterations, the trajectories become easier to mimic for the function approximation and less optimal wrt. the original objective function, until a consensus is found. This algorithm reduces Problem \eqref{eq:joint_pb} to a sequence of trajectory optimization and regression problems, each of which is well-studied with efficient solving method. The complete algorithm is summarized in Algorithm \ref{gtl_algo}. A suitable stopping criterion is $\left\| \Lambda^{k+1} - \Lambda^{k}\right\| \leq \epsilon$, where $\epsilon$ depends on the need of accuracy.

\subsection{Convergence Analysis} \label{sect:conv_anal}

Originally, ADMM was intended to solve convex unconstrained optimization problems, but it has been proven to converge for nonconvex consensus problems \cite{Hong2015, Magnusson2016, Hong2017, Zangwill1969}. \par  \vspace{1.0em}

\begingroup \vspace*{-0.8em}
\removelatexerror
\SetAlgoSkip{}
\LinesNumberedHidden
\SetAlCapHSkip{0.3em}
\begin{algorithm}[H] \label{gtl_algo}
    \Numberline generate N tasks uniformly sampled, $\{\tau_i\}_{i=1}^N \sim U(\mathcal{D}_\tau)$ \\ \vspace{.5em}
    \Numberline initialize $(X^0, U^0, T^0)$ by solving the original trajectory optimization problem for each task in parallel using \eqref{eq:traj_opt} \\
    initialize $\Lambda^0$ to zero \\
    update $W^0$ using the standard regression \eqref{eq:reg_pb}, deduce $Z^0$ \\ \vspace{.5em}
    \Numberline \While{not converged}{
        update $(X^{k+1}, U^{k+1}, T^{k+1})$ by solving the modified trajectory optimization problem in parallel using \eqref{eq:admm_opt} \\
        update $W^{k+1}$ and deduce $Z^{k+1}$ using \eqref{eq:admm_reg} \\[0.23em]
        update $\Lambda^{k+1}$: $\Lambda^{k+1} = \Lambda^{k} + \alpha \left((X^{k+1},\gamma T^{k+1}) - Z^{k+1}\right)$ \\[0.2em]
    }
    \caption{Guided Trajectory Learning}
\end{algorithm} \vspace{1.4em}
\endgroup

\begin{proposition} \label{th:the_prop}
Algorithm \ref{admm_algo} converges to the closest stationary point to $Z^0$, corresponding to a local or global minimum of Problem \eqref{eq:joint_pb_gen}, under these assumptions \cite{Magnusson2016}:
\begin{itemize}
\item The consensus optimization problem \eqref{eq:joint_pb_gen} is feasible.
\item $\forall k \in \mathbb{N}$, $y^k_i$ (resp. $Z^k$) computed at step 2 (resp. step 3) of the algorithm is locally or globally optimal.
\item Let $\mathcal{L}$ denote the set of limit points of the sequence $\{(Y^k,Z^k)\}_{k \in \mathbb{N}}$ and let $(Y^*,Z^*) \in \mathcal{L}$. $(Y^*,Z^*)$ is a regular point, i.e the gradient vectors at $y_i^*$ (resp. $Z^*$) of the set of active constraints of $\mathcal{Y}_i$ (resp. $\mathcal{Z}$) are linearly independent.
\item Let define $L$ such that $\forall y_i \in \mathcal{D}_y, \forall Z \in \mathcal{D}_z$, $f_i$ and $g_i$ have a $L$-Lipschitz continuous gradient. The sequence $\{\rho^k\}_{k \in \mathbb{N}}$ is increasing and either:
    \begin{itemize}
    \item $0 < \alpha \leq 1$ and $\exists k_0 \in \mathbb{N}$ st. $\forall k \geq k_0, \; \rho^k > L$.
    \item $\alpha = 0$ and $\{\rho^k\}_{k \in \mathbb{N}} \longrightarrow +\infty$.
    \end{itemize}
\end{itemize}
\end{proposition} \par \medskip
It is worth noting that updating the penalty factor $\rho^k$ at each iteration is unnecessary for $0 < \alpha \leq 1$, as one can keep it equal any value satisfying the converge assumptions. Andreani has proven that, under the additional assumptions, it converges R-linearly for $\alpha$ small enough and $\rho$ constant \cite{Andreani2007}. Strictly increasing $\rho^k$ makes the converge faster (superlinearly in some cases), but it is impracticable at some point. Indeed, high penalty factor leads to ill-conditioning, making the optimization impossible to solve numerically. \par \medskip
Proposition \ref{th:the_prop} shows that the GTL algorithm can be simplified by setting the dual step size $\alpha$ to 0, which is denoted GTL-0 in the following. It is only guarantee to converge for sequences of penalty factor going to infinity. Otherwise, the price to pay is a non-vanishing reconstruction error. GTL-0 reduces GTL to an instance of the Alternating Direction Penality Method \cite{Magnusson2016} since the multipliers are kept equal to zero. This problem can also be viewed as replacing the reconstruction constraints $X_i = \hat{X}(\tau_i,W), T_i = \hat{T}(\tau_i,W), \, \forall i \in \{0,1,\dots,N\}$ in Problem \eqref{eq:joint_pb} by the penalized reconstruction cost $\rho^k R(X,T,W)$ and solving it in an alternating Gauss-Seidel manner.

\subsection{\fontsize{9.9}{9.9}\selectfont Deconvolution Neural Network as Function Approximation}

\begin{figure*}[!t] \centering
    \includegraphics[trim={0 0.4em 0 -0.5em}, clip, width=\linewidth]{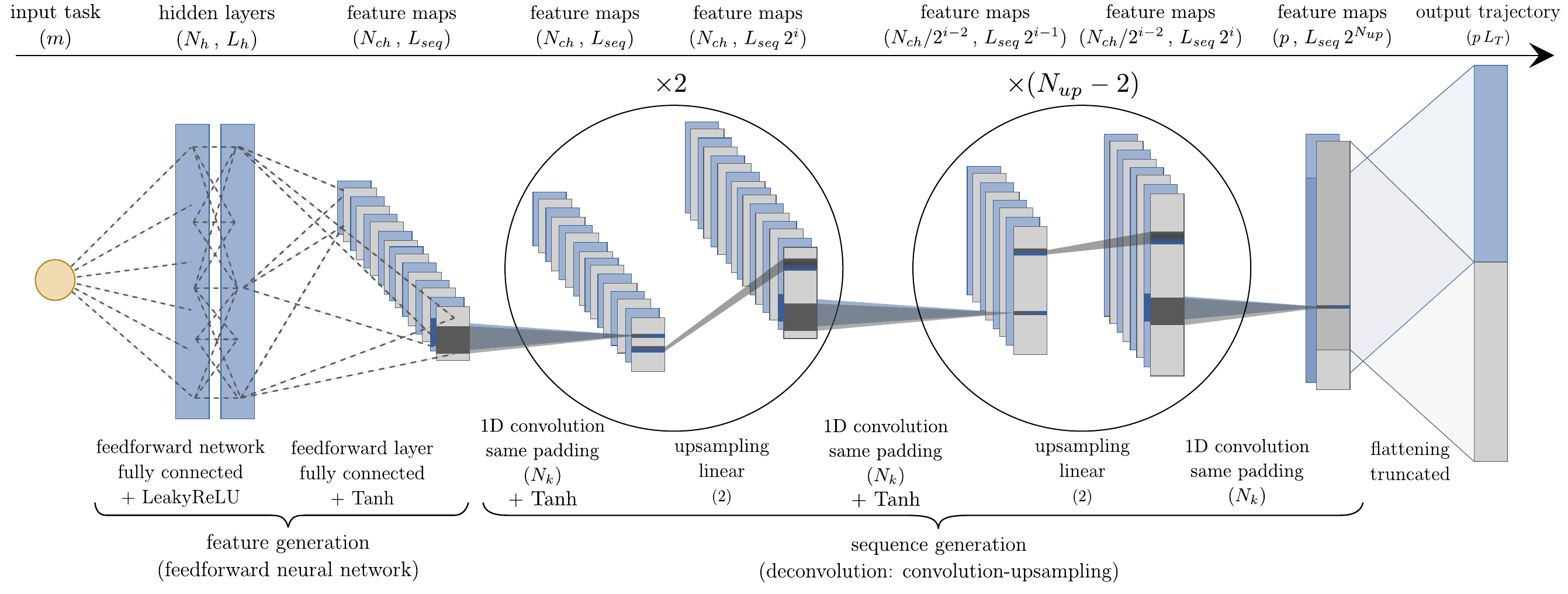}
    \caption{Architecture of the neural network used for learning state sequences. A feedforward network generates low dimensional features and a deconvolution network produces sequences from them. The hyperparameters of the feedforward network are the number of hidden layers $N_{h}$ and their size $L_{h}$. The ones of the deconvolution network are the number of upsampling steps $N_{up}$ and the length of the 1D convolution kernel $N_{k}$. The number of channels $N_{ch}$ and the initial length of the features $L_{seq}$ directly derive from them, $N_{ch} \triangleq p\; 2^{N_{up} - 2}, ~ L_{seq} \triangleq \text{ceil}(L_T / 2^{N_{up}})$. $i$ denotes the index of the upsampling layer. } \vspace{-1.2em}
    \label{fig:deconv_net_archi}
\end{figure*}

We propose to use a deconvolution neural network \cite{Wojna2019, Dong2015} as function approximation since it is especially well-suited to generating multi-dimensional temporal sequences \cite{Tachibana2018}. The architecture is described in Fig \ref{fig:deconv_net_archi}. It combines 1D convolution and upsampling to perform the deconvolution operations, as opposed to the usual transpose convolution that is sensitive to artefacts \cite{Odena2016}. 

\section{EXPERIMENTAL EVALUATION}

\subsection{Experimental Setup}

\subsubsection*{The Medical Exoskeleton Atalante}

It is a crutch-less exoskeleton for people with lower limb disabilities. It is an autonomous device, self-balancing and self-supporting. It has 6 actuated revolute joints on each leg, 
\begin{itemize}
    \item 3 joints for the spherical rotation of the hip,
    \item 1 joint for the flexion of the knee,
    \item 2 joints for the hinge motion of the ankle.
\end{itemize}
It features dimensional adjustments for the thigh and tibia lengths to fit the morphology of the patient. \par \medskip

\subsubsection*{Modeling of the Coupled System Patient-Exoskeleton}

The patient is assumed to be rigidly fastened, thus his mass distribution can be aggregated to the one of the exoskeleton. With this in mind, the system exoskeleton-patient is just a specific type of bipedal robot whose kinematics and dynamics properties are patient-specific. \par \medskip

\subsubsection*{How Trajectories are Generated}

The state sequences must guarantee the periodicity of the gait, accurate impact handling, and stability of the exoskeleton. %
The optimization problem and how to solve it via the Direct Collocation framework is explained thoroughly in \cite{Hereid2016, Grizzle2018, Hereid2018}. 

\begin{figure}[!htbp] \centering
    \includegraphics[trim={0 0.9em 0 0.4em}, width=.93\linewidth]{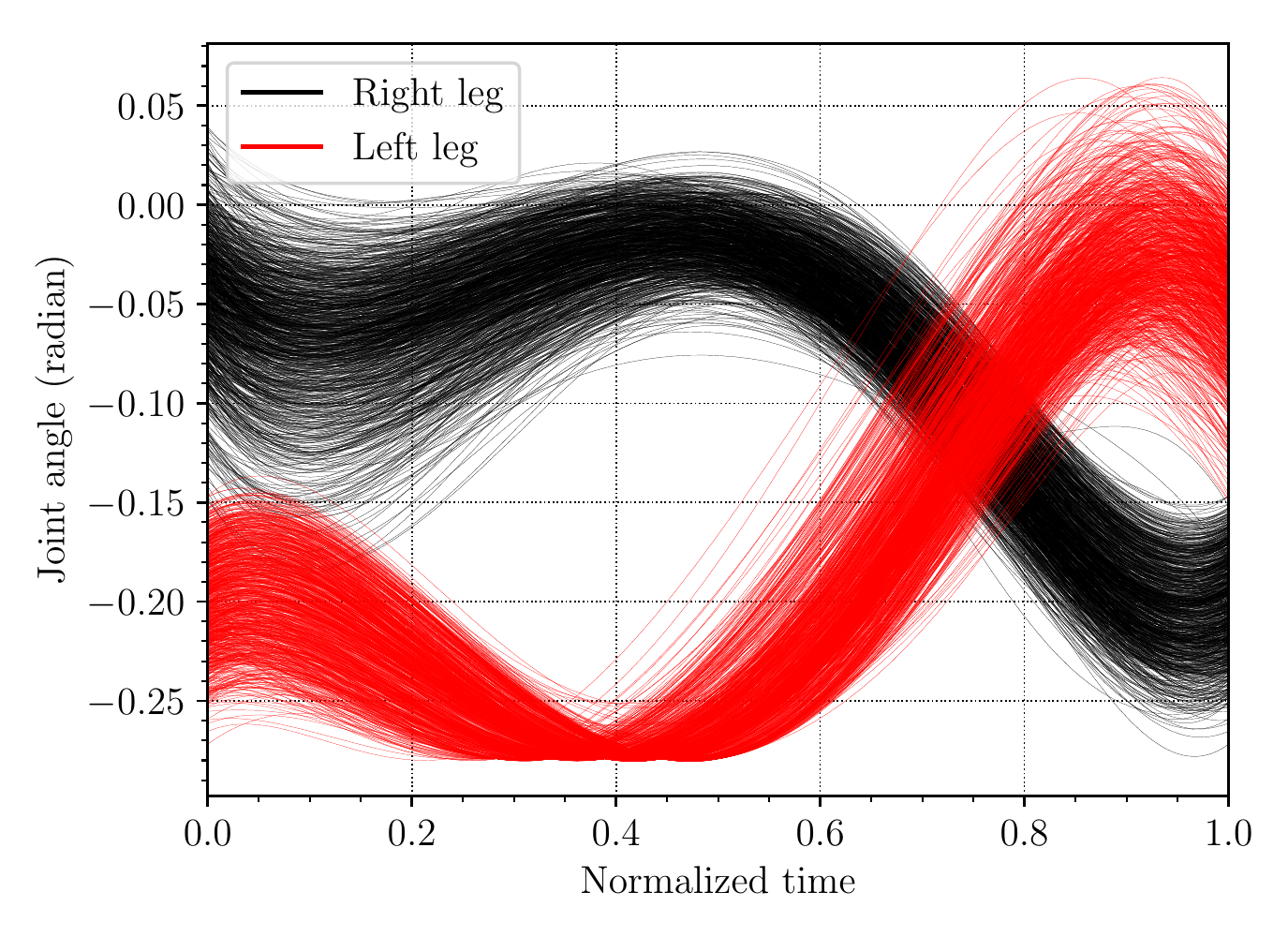}
    \caption{Original trajectories of the ankle joints for uniformly sample tasks.}
    \label{fig:variab_ankle_over_time_gtl} \vspace{-1.5em} 
\end{figure}

\subsection{Training}

\subsubsection*{Learning task}

The objective is to learn flat foot walking trajectories for the exoskeleton Atalante. Fig \ref{fig:variab_ankle_over_time_gtl} reveals their temporal smoothness, supporting the use of the aforementioned deconvolutional network as the natural way to do it. The system is fully-actuated, thereby the state only comprises the positions and velocities of the 12 actuated joints. %
\par \medskip
Standard regression is compared to GTL-0 (cf. last part of Section \ref{sect:conv_anal}) instead of GTL for practical reasons. It halves the required storage space wrt. GTL, but most importantly, it enables to update the sample tasks at each iteration of the algorithm instead of keeping the same set all along. It makes parallelization trivial to implement on clusters, and it reduces overfitting that may occur when the number of sample tasks $N$ is relatively limited. Moreover, its performance can only be worse than GTL, giving a lower bound on the expected performance of GTL. The values of the parameters are summarized in Table \ref{tab:learn_params}. The task space $\mathcal{D}_\tau$ encompasses,
\begin{itemize}
    \item the morphology of the patient: height and weight,
    \item the settings of the exoskeleton: thigh and shank lengths,
    \item some high-level features of the gait: step length and total duration, among many others (12 in total).
\end{itemize} \par \medskip

\begin{table}[!htbp] \vspace{-0.9em}
    \setlength\tabcolsep{0.0pt}
    \caption{Parameters summary} \vspace{-1.3em}
    \subfloat[%
        learning problem \label{tab:learn_params_prob}%
    ]{%
        \begin{minipage}{.32\linewidth} \centering
        \begin{tabularx}{0.95\linewidth}{YYYY}
            \toprule
            \mc{$m$} & \mc{$p$} & \mc{$q$} & \mc{$L_{T}$} \\ 
            \midrule
            16 & 24 & 12 & 200  \\
        \bottomrule
        \end{tabularx}
        \end{minipage}%
    }%
    \subfloat[%
        neural network \label{tab:learn_params_nn}%
    ]{%
        \begin{minipage}{.27\linewidth} \centering
        \begin{tabularx}{0.95\linewidth}{YYY}
            \toprule
            \mc{$N_{h}$} & \mc{$L_{h}$} & \mc{$N_{up}$} \\ 
            \midrule
            1 & 200 & 5  \\  
            \bottomrule
        \end{tabularx}
        \end{minipage}%
    }%
    \subfloat[%
        GTL \label{tab:learn_params_gtl}%
    ]{%
        \begin{minipage}{.40\linewidth} \centering
        \begin{tabularx}{0.95\linewidth}{YYYY}
            \toprule
            \mc{$N$} & \mc{$\gamma$} & \mc{$\rho^k$} & \mc{$\alpha$}\\ 
            \midrule
            70000 & 1.0 & 5.0 & 0.0 \\  
            \bottomrule
        \end{tabularx}
        \end{minipage} 
    } 
    \label{tab:learn_params}
\end{table}

\subsubsection*{Validation criteria}

The controllers of the exoskeletons are tuned in such a way that the maximum tracking error of the joint positions can reach up to 0.01\,rad in the nominal case, which is accurate enough to achieve stable walking. Thus, we assume that a predicted trajectory is stable on the real robot if the maximum absolute difference between a predicted trajectory and the optimal one is not significantly larger than 0.01rad, referred to as norm-inf error in the following. \par \medskip

\subsubsection*{Guided Trajectory Learning}

\begin{figure}[!htbp] \centering
    \includegraphics[trim={0 0.95em 0 -0.8em}, width=.89\linewidth]{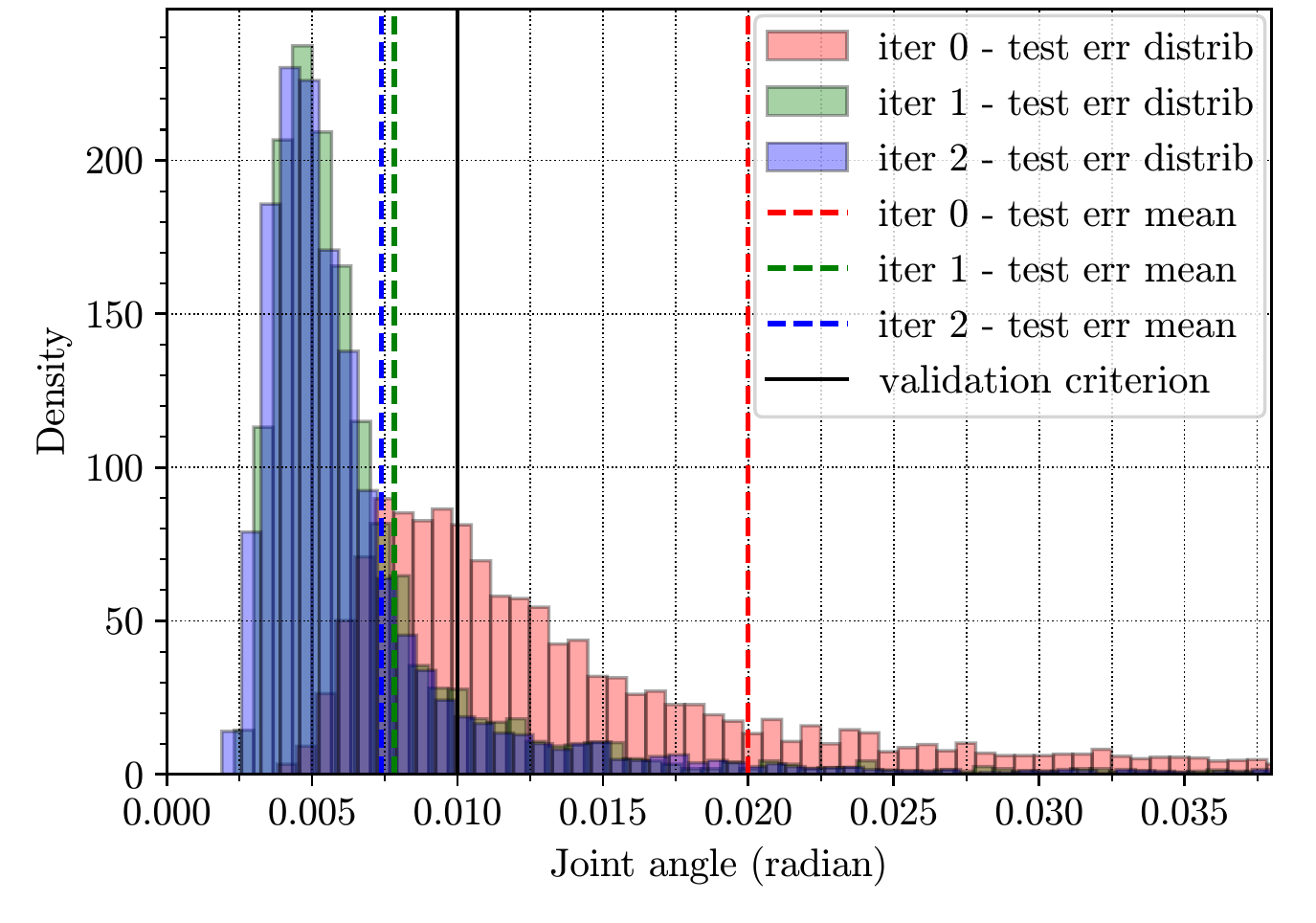}
    \caption{Norm-inf test error distribution over iterations of GTL-0.}
    \label{fig:improve_gtl_pred_err_dist_all_iter_full} \vspace{-1.5em}
\end{figure}

We refer to the initialisation of GTL-0 `iter 0' as Regression since it corresponds to the standard regression method, and we compare it to GTL-0 after convergence at `iter 2'. Their respective accuracy is summarized in Table \ref{tab:test_accuracy}. Unlike Regression, the GTL-0 algorithm shows promising results despite the lack of multipliers. Fig \ref{fig:improve_gtl_pred_err_dist_all_iter_full} shows that, contrary to GTL-0, the error distribution of Regression is very spread and has a long right tail that never really goes to zero. Therefore, a large part of its predictions has a reconstruction error much larger than the maximum acceptable error of 0.01\,rad. Nonetheless, the reconstruction error of GTL-0 does not vanish. This is expected since the residual error is typically handled by the multipliers. It is possible to reduce it further if necessary by increasing the penalty factor $\rho$, at a cost of lowering the conditioning of the optimization problem. \par
\begin{figure}[!htbp] \centering
    \includegraphics[trim={1.5em 0.9em 0 0.8em}, width=.89\linewidth]{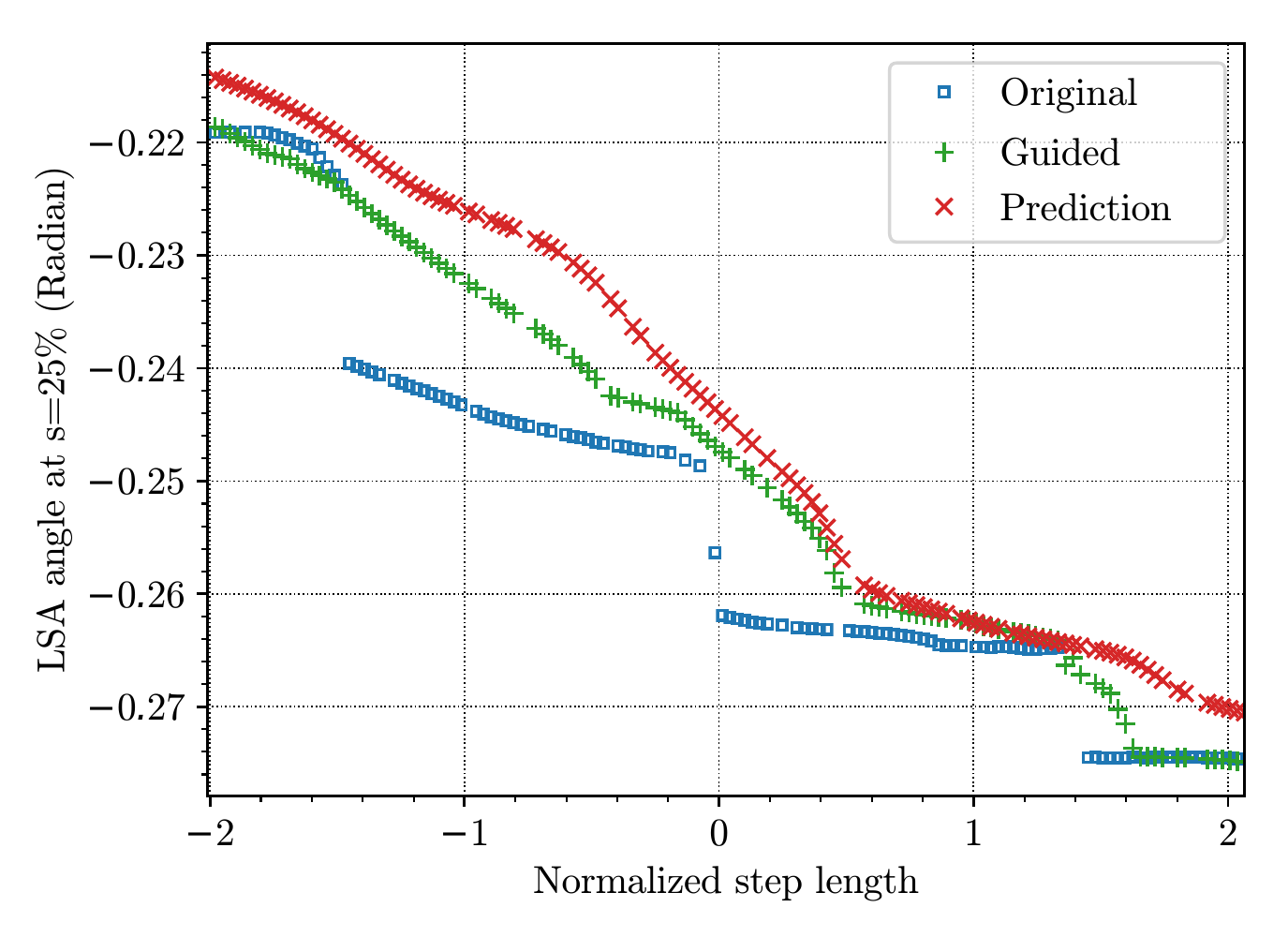}
    \caption{Continuity of the trajectories wrt. the task. It shows the effect of the variation of the step length of the walking gait on the angle of the left ankle joint at 20\% of the step (see Fig \ref{fig:variab_ankle_over_time_gtl}). `Original' denotes the solutions of the original trajectory optimization problem for energy minimization cost. `Guided' and `Prediction' correspond to solutions and predictions of GTL-0.}
    \label{fig:cont_traj_vec_features_all} \vspace{-0.8em}
\end{figure}

The efficiency of GTL-0 can be understood in the light of Fig \ref{fig:cont_traj_vec_features_all}. It reveals several discontinuities for the solutions to the original problem, which are impossible to fit accurately using a continuous function approximation. By contrast, the trajectories generated via GTL-0 are perfectly continuous wrt. the task. Only one iteration of GTL is sufficient the continuity of the solutions, thereby explaining the very fast convergence of the algorithm in only 2 iterations. \par

\begin{table}[!htbp]
    \setlength\tabcolsep{0.0pt}
    \renewcommand{\arraystretch}{1.2}
    \caption{Testing accuracy in norm-inf}
    \begin{tabularx}{\linewidth}{YYYYY}
        \toprule
        Algorithm & Mean (rad) & Mode (rad) & \mc{$> 0.01$\,rad} & \mc{$> 0.015$\,rad} \\ 
        \midrule
        Regression & $2.01 \times 10^{-2}$ & $8.16 \times 10^{-3}$ & $50.3\%$ & $16.1\%$ \\
        GTL-0 & $7.43 \times 10^{-3}$ & $4.25 \times 10^{-3}$ & $10.5\%$ & $4.46\%$ \\
        \bottomrule
    \end{tabularx} \vspace{-1.16em}
    \label{tab:test_accuracy}
\end{table}

\subsection{Validation in reality on able-bodied people}

\begin{figure}[!htbp] \centering \vspace{-0.2em}
    \includegraphics[trim={0 4.4em 0 1.2em}, clip, width=0.96\linewidth]{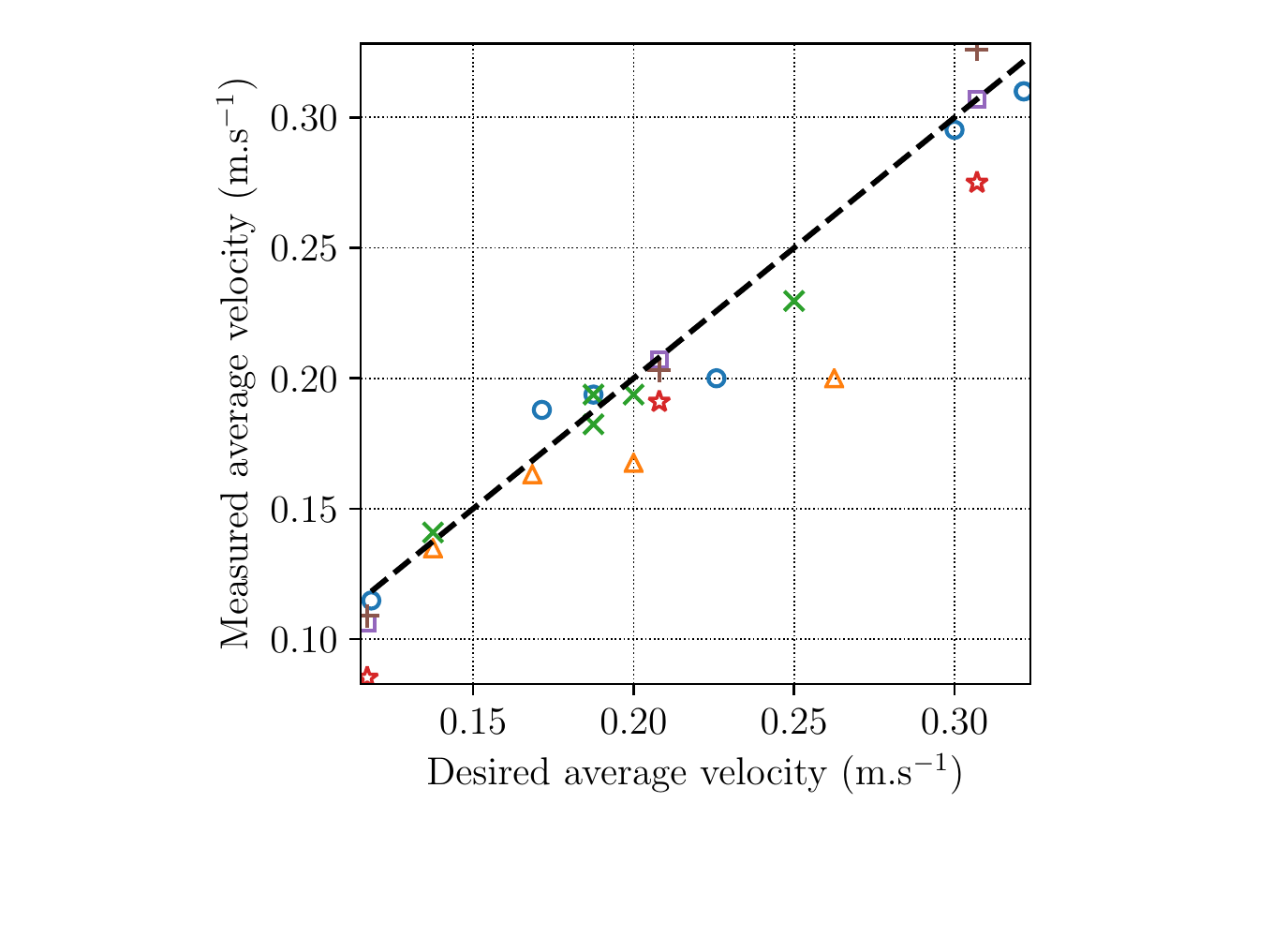}
    \caption{From simulation to reality. Comparison between the desired and achieved average velocity on 6 valid people with a different morphology. Each pair marker-color corresponds to one patient. %
    } \vspace{-1.0em}
    \label{fig:valid_speed_patient}
\end{figure}

\noindent We have evaluated our ability to control the average velocity of the exoskeleton. The desired average velocity is determined by the combination of desired step length and duration. Note that data are only available for GTL-0, since most predictions were unstable on the real robot using the standard regression. In the case of GTL-0, the vast majority of them were stable, and restricting the ranges of the desired step length and duration to 90\% during inference lead to stable gaits only. Fig \ref{fig:valid_speed_patient} shows that the measured velocities are close to the desired ones for every patient.

\section{CONCLUSION AND FUTURE WORK}

In this work, we present a novel algorithm called GTL that learns a function approximation of the solutions to a trajectory optimization problem over a task space. Accurate and reliable predictions are ensured by simultaneously training the function approximation and adapting the trajectory optimization problem such that its solutions can be perfectly fitted by the function approximation and satisfy the constraints concurrently. It results in a consensus optimization problem that we solve iteratively via ADMM. We demonstrate its efficiency on flat-foot walking with the exoskeleton Atalante. \par
We believe that our method offers a new scope of applications, such as reinforcement learning, perturbation recovery, or path replanning. Enabling adaption of the architecture of the neural network itself to further improve its efficiency and usability is an exciting direction for future work.


\clearpage
\bstctlcite{BSTcontrol}
\bibliographystyle{IEEEtran}
\bibliography{IEEEabrv,bibliography}

\end{document}


\clubpenalty=9996
\widowpenalty=9999 
\brokenpenalty=4991
\predisplaypenalty=10000
\postdisplaypenalty=1549
\displaywidowpenalty=1602

\maketitle


\section{From continuous to discrete problem}

In this paper, it is implicitly assumed that the task space is a finite set of point rather than a continuous space, for the sake simplicity. The objective of this section is to show under which conditions the theory presented in this paper remains valid. It consists in computing error bounds resulting from the discretization of the problem using a finite set of $N$ trajectory optimization problems to approximate the solutions of a nonconvex optimization problem over a continuous task space. \par \medskip
Combining trajectory optimization and trajectory learning can be understood as optimizing a function parameterized by $W \in \mathbb{R}^n$ such that
\begin{equation} \label{eq:ref_pb_orig}
    (U^*,W^*) = \underset{u(\cdot),W}{\argmin}\mkern+05mu \int_{\substack{\phantom{1} \\ \tau \in \mathcal{D}_\tau}} \mkern-20mu L(\hat{X}(\tau,W),u(\tau)) \, d\tau
\end{equation} \vspace{-1.4em}
\begin{align*}
    \text{st. } \forall \tau \in \mathcal{D}_\tau, \quad
    & (\hat{X}(\tau,W),u(\tau),\hat{T}(\tau,W)) \in \mathcal{C}_\tau
\end{align*}

Equivalently, it can be viewed as the following consensus optimization problem
\begin{equation} \label{eq:ref_pb_cons}
    (X^*,U^*,T^*,W^*) = \underset{x(\cdot),u(\cdot),T(\cdot),W}{\argmin}\mkern+05mu \int_{\substack{\phantom{1} \\ \tau \in \mathcal{D}_\tau}} \mkern-20mu L(x(\tau),u(\tau)) \, d\tau
\end{equation} \vspace{-1.4em}
\begin{align*}
    \text{st. } \forall \tau \in \mathcal{D}_\tau, \quad
    & (x(\tau),u(\tau),T(\tau)) \in \mathcal{C}_\tau \\
    & x(\tau) = \hat{X}(\tau,W), \, T(\tau) = \hat{T}(\tau,W)
\end{align*}

The total trajectory optimization cost $L$ is a continuously differentiable function and the task space is a compact. Thus, the integral can be approximated using Monte-Carlo integration, which can be stated as follow
\begin{equation*}
     I_N = \frac{V}{N} \sum_{i=1}^{N} L(x(\tau_i),u(\tau_i))
\end{equation*}
where $N$ is the number of samples, $V$ is the volume of $\mathcal{D}_\tau$, $\tau_i$ are sample uniformly on $\mathcal{D}_\tau$. \par \medskip
Then, it can be proven that

\begingroup
\vspace{-0.8em}
\centering
\setlength\tabcolsep{0.0pt}
\begin{tabularx}{0.95\linewidth}{YY}
\begin{equation*}
    \lim_{N \to \infty} I_N = \int_{\substack{\phantom{1} \\ \tau \in \mathcal{D}_\tau}} \mkern-20mu L(x(\tau),u(\tau))
\end{equation*}
& \vspace{-0.9em}
\begin{equation*}
    \text{Var}(I_N) = V^2 \frac{\text{Var}(f)}{N}
\end{equation*}
\end{tabularx}
\endgroup

where $\text{Var}(f)$ is defined by
\begin{equation*} \vspace{-1.0em}
     \text{Var}(f) = \frac{1}{N-1} \sum_{i=1}^{N} (f(x(\tau_i),u(\tau_i)) - I_N)^2
\end{equation*}

Therefore, the optimal solutions of the continuous problem can be
approximated at $O(N^{-1/2})$ in terms of optimality by \vspace{-0.2em}
\begin{equation*}
    (X^*,U^*,T^*,W^*) = \underset{x(\cdot),u(\cdot),T(\cdot),W}{\argmin}\mkern+05mu \sum_{i=1}^{N} L(x(\tau_i),u(\tau_i)) \, d\tau
\end{equation*}
\begin{equation*}
    \text{st. } \forall \tau \in \mathcal{D}_\tau, \quad
    (x(\tau),u(\tau),T(\tau)) \in \mathcal{Z}_{\tau,W}
\end{equation*}
where $\mathcal{Z}_{\tau,W}$ is defined by
\begin{align*}
   \mathcal{Z}_{\tau,W} = \;
   & \{(x,u,T) \in \mathcal{C}_\tau | \, x(\tau) = \hat{X}(\tau,W), \, T(\tau) = \hat{T}(\tau,W)\} \\
   & \{(x,u,T) \in \mathcal{D}_X \times \mathcal{D}_U \times \mathcal{D}_T | \, c(\tau,x,u,T,W) \leq 0\}
\end{align*}

The constraints still hold continuously over the task space $\mathcal{D}_\tau$ and it is not straightforward to discretize them. To An error bound on the constraints violation $\mu$ induced by the discretization must be estimated first, namely
\begin{equation*}
    \forall \tau \in \mathcal{D}_\tau, \; \inf_{u \in \mathcal{D}_U} c_i(\tau,\hat{X}(\tau,W^*),u,\hat{T}(\tau,W),W^*) \leq \mu
\end{equation*}
where $c_i$ is the $i$-th component of $c$. \par \medskip
In addition, according to the Mean Value Theorem over a compact set, we have
\begin{align*}
    \forall \tau, \tau' \in \mathcal{D}_\tau, \;
    & \left\| \hat{X}(\tau,W) - \hat{X}(\tau',W) \right\| \leq
    K \left\| \tau - \tau' \right\| \\
    & \left\| \hat{T}(\tau,W) - \hat{T}(\tau',W) \right\| \leq
    L \left\| \tau - \tau' \right\|
\end{align*}
where $k_i=\max_{\tau \in \mathcal{D}_\tau} \left\| \nabla_\tau \hat{X}_i(\tau,u,T,W) \right\|$, and $l_i=\max_{\tau \in \mathcal{D}_\tau} \left\| \nabla_\tau \hat{T}_i(\tau,u,T,W) \right\|$. \par \medskip
Similarly, one can compute bounds on the constraints
\begin{equation*}
    \forall z, z' \in \mathcal{D}_\tau \times \mathcal{D}_x, \;
    \lvert c_i(z,u,W) - c_i(z',u,W)\rvert \leq
    m_i \left\| z - z' \right\|
\end{equation*}
where $z=(\tau,X(\tau,W),T(\tau,W))$, and $m_i=\max_{\tau \in \mathcal{D}_\tau}\lvert \nabla_z c_i(z,u,T,W) \rvert$. \par \medskip
Combining those two inequalities, it yields
\begin{multline} \label{eq:apdx_inq_proof}
    \forall \tau, \tau' \in \mathcal{D}_\tau, \;
    c_i(\hat{X}(\tau,W),u,\hat{T}(\tau,W),W) - \\ c_i(\hat{X}(\tau',W),u,\hat{T}(\tau',W),W)
    \leq m_i (1 + K + L) \left\| \tau - \tau' \right\|
\end{multline}

Let's consider a set of task points $\{\tau_i\}_{i \leq N}$ such that
\begin{equation*}
    \exists \epsilon > 0, \; \forall \tau \in \mathcal{D}_\tau, \; \max_{\tau_i \in \mathcal{G}_\tau}\left\|\tau - \tau_i\right\| \leq \epsilon
\end{equation*}

Since $\forall \tau \in \mathcal{G}_\tau, \, \inf_{u \in \mathcal{D}_U} c_i(\hat{X}(\tau,W),u,\hat{T}(\tau,W),W) \leq 0$, it follows from \eqref{eq:apdx_inq_proof}
\begin{multline*}
    \forall \tau \in \mathcal{D}_\tau, \; \inf_{u \in \mathcal{D}_U} c_i(\tau,\hat{X}(\tau,W^*),u,\hat{T}(\tau,W),W^*) \\
    \leq M_i (1 + K + L) \epsilon
\end{multline*}

A tighter bound on the constraint violations can be obtained by defining $M_i$ and $K$ as local maxima instead of global ones. It follows
\begin{align*}
    m_i = & \max_{\left\|\tau - \tau'\right\| \leq \epsilon}\lvert \nabla_\tau c_i(\tau',\hat{X}(\tau',W),u,\hat{T}(\tau',W),W) \rvert \\
    k_i = & \max_{\left\|\tau - \tau'\right\| \leq \epsilon} \left\| \nabla_\tau \hat{X}_i(\tau,u,T,W) \right\|
\\
    l_i = & \max_{\left\|\tau - \tau'\right\| \leq \epsilon} \left\| \nabla_\tau \hat{T}_i(\tau,u,T,W) \right\|
\end{align*}

Assuming that $c,\hat{X},\hat{T}$ are twice continuously differentiable, it is possible to get rid of the $\max$ and use instead of global maxima of the second derivatives. \par \medskip
Finally, one can compute an probabilistic estimate of $\epsilon$. The smallest value for $\epsilon$ is given by
\begin{equation*}
    \epsilon_N = \max_{\tau_i \in \mathcal{G}_\tau} { \min_{\tau_j \in \mathcal{G}_\tau}\left\|\tau_i - \tau_j\right\|}
\end{equation*}
where $\tau_i, \tau_j$ are sample uniformly on $\mathcal{D}_\tau$. \par \medskip
According to the Extreme Value Theorem, if there exist two sequences $a_n>0, b_n \in \mathbb{R}$ such that
\begin{equation*}
    \mathbb{P}\left(\frac{M_n - b_n}{a_n} \leq z\right) \xrightarrow[n \to \infty]{} G(z)
\end{equation*}
where $M_n=\max{(X_1,X_2,\dots,X_N)}$ and $X_1,X_2,\dots,X_N$ is a sequence of independent and identically distributed random variables. \par
Then $G(z)$ has to be a Gumbel distribution, i.e there exist $a,b \in \mathbb{R}$ such that
\begin{equation*}
    G(z) = \exp{\left(-\exp{\left(-\frac{x-b}{a}\right)}\right)}
\end{equation*}
Suitable values of $a_n, b_n, a, b$ has been estimated by several authors \cite{Dette1990, Gyorfi2019}. It particular,
\begin{equation*}
    \mathbb{P}\left(N \omega_m \epsilon_N^m - \log{N} \leq t\right)  \xrightarrow[n \to \infty]{} e^{e^{-x}}
\end{equation*}
where $m$ is the dimension of the task space, and $\omega_m = \pi ^ {m/2} / \Gamma(1+m/2)$ is the volume of the unit sphere in $\mathbb{R}^m$.\par \medskip
According to Jensen's inequality and using the fact $\mathbb{E}(X) \leq \mathbb{E}(X^+)$, an upper bound of the expectation of $\epsilon_N$ can be derived
\begin{equation*}
    \mathbb{E}(\epsilon_N) \leq \beta \left(\frac{\log{N}}{N}\right)^{1/m}
\end{equation*}
which turns out to be a very thigh bound for $N$ large. \par \medskip
Only the case of uniform distribution of sample tasks was considered here for simplicity, but it generalizes to other distribution almost effortlessly. As one could expect, it shows that the solution of the discrete problem converges asymptotically to the solution of the continuous one. The actual limitation of this discretization of the problem is the constraint violation, such decreases slowly with the number of samples. Moreover, notice that it the dynamics equation $\dot{x} = f_{\tau} (x,u)$ is violated as much as any other constraints that depends on the task to solve. It means that the trajectory will not be properly executed on the system, leading to the quadratic compounding of errors over time. This is a common issue related to transfer learning. It has similar effects than model uncertainties and approximations. Many traditional nonlinear control strategies has been designed to handle them. \par
It is worthy to note that the upper bounds estimate on the constraint violations could be used to enable adaptive number of sample task $N$.

\section{Existence of a solution}

\begin{lemma} \label{th:the_lemma}
If the task space $\mathcal{D}_\tau$ is a compact set, the constraints $c_{in}, c_{eq}$ of the trajectory optimization problem are continuous wrt. the task $\tau$, and $\forall \tau \in \mathcal{D}_\tau$ the feasible set $\mathcal{C}_\tau$ is non-empty, then there exists a continuous closed map $\mathcal{M}$ from the task space to the feasible set, i.e. $\mathcal{M}:\tau \in \mathcal{D}_\tau \mapsto (x(\tau),u(\tau),T(\tau) \in \mathcal{C}_\tau$.
\end{lemma} \par \medskip

Formally, the assumption of existence of a solution can be formulated as
\begin{align*}
    \mathcal{F} = \{ & x:\mathcal{D}_\tau \mapsto \mathcal{D}_X, T:\mathcal{D}_\tau \mapsto \mathcal{D}_T \, | \;
    \exists \, u:\mathcal{D}_\tau \mapsto \mathcal{D}_U  \\
    &\text{s.t. } \forall \tau \in \mathcal{D}_\tau, \; (x(\tau),u(\tau),T(\tau)) \in C_{\tau} \} \neq \emptyset
\end{align*}

The discrete version of it follows directly from the previous results,
\begin{equation*}
     \lim_{n \to \infty} \prod_{i \in \{0,1,\dots,n\}} \mkern-20mu \mathcal{C}_{\tau_i} \; \cap \, \mathcal{Z} = \mathcal{F} \neq \emptyset
\end{equation*}

The Universal Approximation Theorem \cite{Zhou2017, Hornik1991, Hanin2017} can be expressed as follow: let $\phi:\mathbb{R}\to\mathbb{R}$ be a nonconstant bounded and continuous activation function. Let $I_m$ denote the m-dimensional unit hypercube $[0,1]^m$. The space of real-valued continuous functions on $I_m$ is denoted by $C(I_m)$. Then, given any $\epsilon>0$ and any function $f \in C(I_m)$, there exist an integer $N$, real constants $v_i,b_i \in \mathbb{R}$ and real vectors $w_i \in \mathbb{R}^m$ for $i=1,\ldots,N$, such that we may define
\begin{equation*}
     \hat{f}(x) = \sum_{i=1}^{N} v_i \varphi \left( w_i^T x + b_i \right)
\end{equation*}
as an approximate of the function $f$ such that
\begin{equation*}
     \forall x \in I_m, \quad \left\| \hat{f}(x) - f (x) \right\| < \epsilon
\end{equation*}

It means that functions of the form of $\hat{f}$ are dense in $C(I_m)$. Therefore, any function $f \in \mathcal{F}$ can be approximate arbitrary well by a multilayer feedfoward neural network with usual activation function, providing that the number of parameters $n$ st. $W \in \mathbb{R}^n$ is high enough. \par \medskip
According to Lemma \ref{th:the_lemma} and the Universal Approximation Theorem, there exists a function approximation with bounded expressive power for which the consensus optimization problem is feasible. Consequently, the trajectory optimization problem can be adapted in such a way that the function approximation fits perfectly its solutions over the whole span of the task space. Choosing a function approximation with a higher expressive power does not affect the reconstruction error, but the trajectories will get closer to the solutions of the original trajectory optimization problem, to perfectly match at the limit. \par \medskip
Note that there is a major difference between Problems \eqref{eq:ref_pb_orig} and \eqref{eq:ref_pb_cons} when the problem is infeasible but close to being. Indeed, using a numerical solver would fail to solve \eqref{eq:ref_pb_orig} and stop at an infeasible point. At this point, there is no way to make sure that the infeasibility is actually minimized. On the contrary, each subproblem in step 2 and 3 of the GTL algorithm are still likely to converge. If so, $X^*,Z^*$ converge despite that $\lambda^*$ does not. As a result, the stopping criterion of the algorithm must not involve $\lambda^*$. In this case, the derivative of the mean of the norm-inf$^2$ of the reconstruction error no longer decreases, that is $\mathbb{E}(\left\|X^k - Z^k\right\|_\infty) - \mathbb{E}(\left\|X^{k-1} - Z^{k-1}\right\|_\infty) \leq \epsilon$. After convergence, $X^*,Z^*$ are much that it minimizes as much as possible the reconstruction error $X^* - Z^*$. Therefore, it gives a gradient to assess if changing the architecture of the network of the number of fitting parameters $n$ goes the right way to reduce the infeasibility.

\section{Extension to Parametric Programming}

Unlike Nonlinear Programming (NLP), Parametric programming aims at solving an optimization problem wrt. some variables as a function to some others. In general, it can be formulated as follow
\begin{equation} \label{eq:ref_pb_params}
    \forall \tau \in \mathcal{D}_\tau \subset \mathbb{R}^m, \quad X^*(\tau) = \argmin_{x \in \mathcal{C}_\tau \subset \mathbb{R}^p}\mkern+05mu f(x,\tau)
\end{equation}
where $x \in \mathbb{R}^p$ is the optimization variable, $\tau$ are the parameters, $f(x,\tau)$ is the objective function, and $\mathcal{C}_\tau$ denotes the feasibility set that depends on $\tau$. The set $\mathcal{D}_\tau$ is generally referred to as parameter space. \par \medskip
This problem can be solved numerically through NLP for any fixed parameter $\tau$. This approach enables to compute the optimal solutions for a finite collection of parameters $\{\tau_i\}_{i \leq N}$, but it does not extended naturally to any value of $\tau$ over a continuous parameter space. To solve this kind of problem,

Parametric programming must to reduced to classic NLP to be solved. We propose to do so by introducing an extra function $\hat{X} : \mathbb{R}^m \times \mathbb{R}^n \mapsto \mathbb{R}^p$ such that
\begin{gather} \label{eq:ref_pb_params_nlp}
    \forall \tau \in \mathcal{D}_\tau, \quad W^* = \argmin_{W \in \mathbb{R}^n}\mkern+05mu f(\hat{X}(\tau, W),\tau) \\
    \text{st. } \hat{X}(\tau, W) \in \mathcal{C}_\tau \nonumber
\end{gather}
where $\hat{X}$ is referred to as function approximation in the following. Then, the solution of Problem \eqref{eq:ref_pb_params} can be approximated by
\begin{equation*}
    \forall \tau \in \mathcal{D}_\tau, \quad \widetilde{X}(\tau) = \hat{X}(\tau, W^*)
\end{equation*}

This approach has several limitations. Firstly, the optimal solution $\widetilde{X}$ of Problem \eqref{eq:ref_pb_params_nlp} only matches the solution $X^*$ of Problem \eqref{eq:ref_pb_params} when the expressive power of the function approximation $\hat{X}(\tau, W),\tau)$ is high enough so that one can find a suitable value of $W \in \mathbb{R}^n$ that can fit any solution of \eqref{eq:ref_pb_params}, namely,
\begin{equation} \label{eq:ref_pb_params}
    \exists \; W \in \mathbb{R}^n, \text{ st. }  \forall \tau \in \mathcal{D}_\tau, \; \hat{X}(\tau, W) = \argmin_{x \in \mathcal{C}_\tau \subset \mathbb{R}^p}\mkern+05mu f(x,\tau)
\end{equation}

In general, such function does not exist and therefore only an approximation is available. Nevertheless, this solution is guarantee to be feasible for any parameter $\tau$, which is what matters in most applications. Indeed, the cost function is usually used to give information to the solver in which direction solutions with suitable properties lie, but finding the solution that is the global minimum of that cost is irrelevant. For instance, the cost function for bipedal walking is the total energy consumption of the system. \par \medskip
The second limitation of this approach is that solving Problem \eqref{eq:ref_pb_params_nlp} requires to compute the solution of an infinity of NLP simultaneously, which is as much impracticable as Problem \eqref{eq:ref_pb_params}. The GTL algorithm is an iterative algorithm that compute an approximate solution of Problem \eqref{eq:ref_pb_params}, whose the degree of approximation can be tuned via the size $N$ of the finite collection of randomly sample parameters $\{\tau_i\}_{i \leq N}$ at which the NLP subproblem is solved. Brought together, it enables to compute the solution of Problem \eqref{eq:ref_pb_params}, providing that the expressive power of the function approximation and the number of sample parameters are high enough.


\bstctlcite{BSTcontrol}
\bibliographystyle{IEEEtran}
\bibliography{IEEEabrv,bibliography}